\documentclass[11pt,a4paper]{article}
\usepackage[hyperref]{acl2020}
\usepackage{hyperref}
\usepackage{times}
\usepackage{latexsym}
\usepackage{graphicx}
\usepackage{arydshln}
\usepackage{adjustbox}
\usepackage{booktabs,tabularx, colortbl}
\usepackage{caption}
\usepackage[fleqn]{amsmath}
\usepackage[noabbrev]{cleveref}
\usepackage{textcomp}

\usepackage{dingbat}
\usepackage{microtype}

\aclfinalcopy 
\def\aclpaperid{2374} 

\definecolor{mypurple}{RGB}{133,3,133}
\definecolor{mygreen}{RGB}{57,118,29}
\definecolor{IndianRed1}{RGB}{255,36,36}
\definecolor{MediumOrchid}{RGB}{170,63,208}
\definecolor{DeepSkyBlue1}{RGB}{0,178,238}
\definecolor{Orange}{RGB}{238,140,0}

\title{Knowledge Graph-Augmented Abstractive Summarization \\with Semantic-Driven Cloze Reward}

\author{Luyang Huang$^{1}$ \quad Lingfei Wu$^{2}$ \quad  {\rm and} \quad Lu Wang$^{1}$\\
  $^{1}$Khoury College of Computer Sciences, Northeastern University, Boston, MA 02115\\
  $^{2}$IBM Research AI, IBM T.J. Watson Research Center, Yorktown Heights, NY 10598 \\
  $^{1}${\tt luyang.huang96@gmail.com, luwang@ccs.neu.edu} \\
  $^{2}${\tt wuli@us.ibm.com} \\}

\date{}

\begin{document}
\maketitle
\begin{abstract}

Sequence-to-sequence models for abstractive summarization have been studied extensively, yet the generated summaries commonly suffer from fabricated content, and are often found to be near-extractive.
We argue that, to address these issues, the summarizer should acquire semantic interpretation over input, e.g., via structured representation, to allow the generation of more informative summaries. 
In this paper, we present \textbf{\textsc{ASGARD}}, a novel framework for Abstractive Summarization with Graph-Augmentation and semantic-driven RewarD. 
We propose the use of {\it dual encoders}---a sequential document encoder and a graph-structured encoder---to maintain the global context and local characteristics of entities, complementing each other. 
We further design {\it a reward based on a multiple choice cloze test} to drive the model to better capture entity interactions. 
Results show that our models produce significantly higher ROUGE scores than a variant without knowledge graph as input on both New York Times and CNN/Daily Mail datasets. We also obtain better or comparable performance compared to systems that are fine-tuned from large pretrained language models. 
Human judges further rate our model outputs as more informative and containing fewer unfaithful errors. 
\end{abstract}

\section{Introduction}

\begin{figure}[t]
    
    \def\arraystretch{1.5}
    \bgroup
    \def\arraystretch{1.5}
    \setlength{\arrayrulewidth}{1pt}
	\fontsize{9}{11}\selectfont
    
	\setlength{\tabcolsep}{0.8mm}
	\hspace{-2mm}
	\includegraphics[width=\columnwidth,trim=0 0 0cm 0, clip]{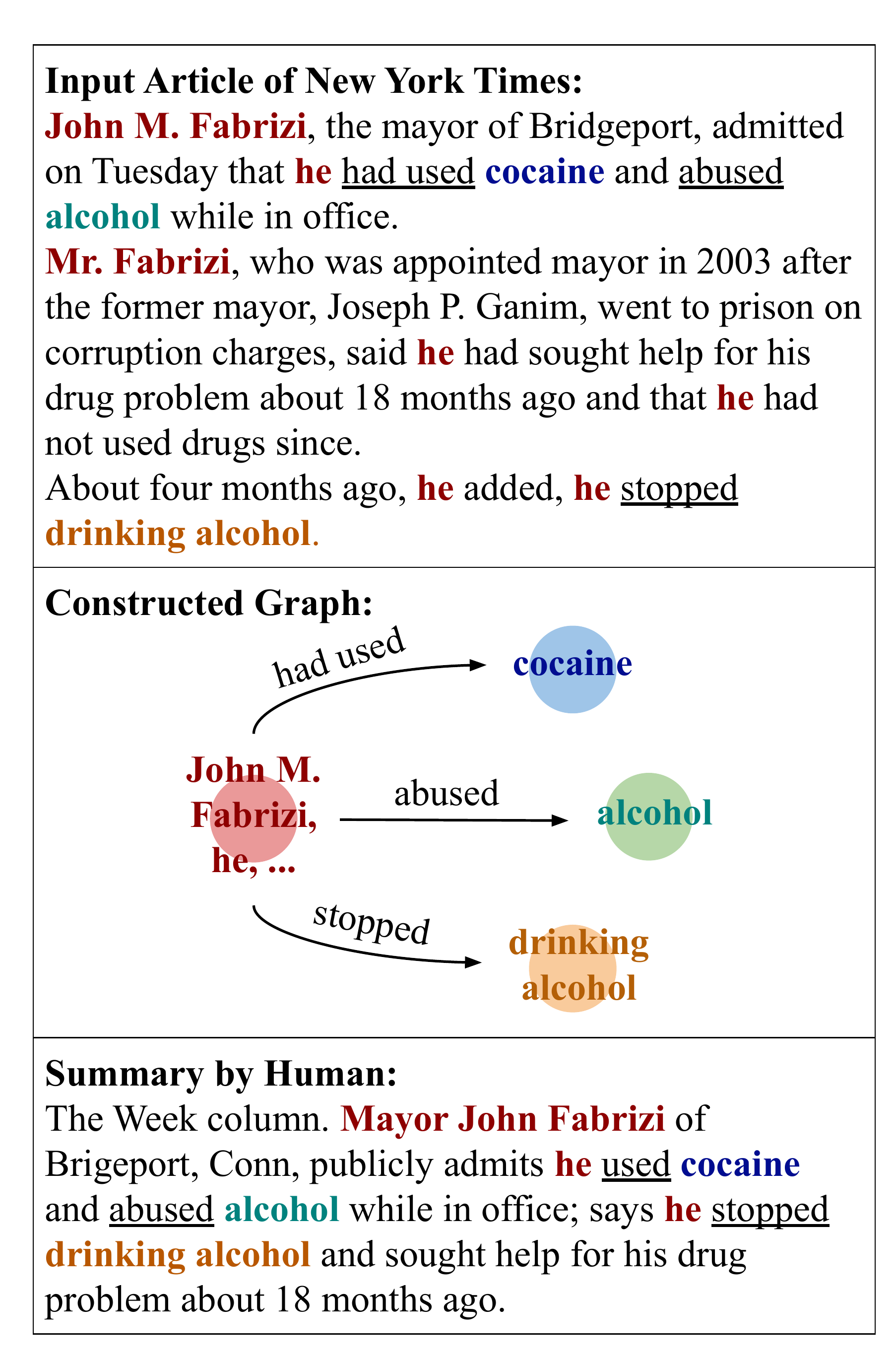}
	\vspace{-3mm}
    \caption{Sample knowledge graph constructed from an article snippet. The graph localizes relevant information for entities (color coded, e.g. ``{\it John M. Fabrizi}'') or events (underlined) and provides global context.
} 
\label{fig:intro}
\egroup
\vspace{-3mm}
\end{figure}

Abstractive summarization aims to produce concise and informative summaries with the goal of promoting efficient information consumption and knowledge acquisition~\cite{luhn1958automatic}. 
Significant progress has been made in this area by designing sequence-to-sequence-based neural models for single-document abstractive summarization~\cite{gehrmann-etal-2018-bottom,j.2018generating,liu-lapata-2019-text}. 
However, due to the limitations of model structure and word prediction-based learning objectives, these models frequently produce unfaithful content~\cite{cao2017faithful} and near-extractive summaries~\cite{see-etal-2017-get,kryscinski2018improving}. These observations suggest that existing models lack semantic interpretation over the input, which is critical for summarization.

We argue that the generation of informative and succinct abstracts requires structured representation to facilitate the connection of relevant subjects, and the preservation of global context, e.g. entity interactions and topic flows. Take Fig.~\ref{fig:intro} as an example. Complex events related with the same entity may span multiple sentences, making it challenging for existing sequential models to capture. A graph representation, on the contrary, produces a structured summary and highlights the proximity of relevant concepts. 

To this end, we present \textbf{\textsc{ASGARD}}, a framework for Abstractive Summarization with Graph-Augmentation and semantic-driven RewarD.\footnote{Our code is available at \href{https://github.com/luyang-huang96/GraphAugmentedSum}{https://github.com/luyang-huang96/GraphAugmentedSum}.} 
Under the encoder-decoder framework, {\it we enhance the regular document encoder with a separate graph-structured encoder to maintain the global context and local characteristics of entities} by using the outputs from an open information extraction (OpenIE) system.

Specifically, we experiment with two graph variants, one mainly capturing entities' document-level interactions and the other reflecting such interactions within each paragraph plus topic shifts across paragraphs. 
Both graphs can capture interactions among entities that are positioned far from one another in the document and significantly reduce redundancy, as shown in Fig.~\ref{fig:intro}. 
The document encoder and the graph encoder then cooperate during abstract generation, wherein the model is trained to identify salient content by aligning graphs with human summaries. Though structured representation has been studied before for summarization~\cite{fernandes2018structured}, to the best of our knowledge, we are the first to utilize graph neural networks to {\it explicitly} encode entity-centered information for abstractive summary generation. 

Moreover, we propose {\it a novel multi-choice cloze reward to drive the model to acquire semantic understanding over the input}. Concretely, we design cloze questions by removing {\it pairwise} entities that are connected with a predicate or co-occur in a human summary sentence, whereas prior work only considers single entities to construct questions~\cite{eyal-etal-2019-question}. 
In tandem with our graph encoding of knowledge, the cloze reward further facilitates the acquisition of global entity interactions with reinforcement learning.

We carry out automatic and human evaluations on popular summarization datasets. 
Models based on \textsc{ASGARD} yield significantly better ROUGE scores~\cite{Lin:2003:AES:1073445.1073465} than a variant without access to the knowledge graph on two popular news summarization datasets, New York Times corpus and CNN/Daily Mail dataset. Moreover, \textsc{ASGARD} models attain performance better than or comparable to others that are fine-tuned from large pretrained language models, including BERTSum~\cite{liu-lapata-2019-text}, UniLM~\cite{NIPS2019_9464}, and BART~\cite{lewis2019bart}. 
Human judges further confirm that our models generate more informative summaries with less unfaithful errors than their counterparts without the graph encoder. 
Importantly, we find that automatic evaluation metrics only weakly correlate with these errors, implying that new evaluation methods are needed to better gauge summary quality. 

The rest of the paper is organized as follows. We describe related work in the next section (\S~\ref{sec:related}). We then discuss the knowledge graph construction in \S~\ref{sec:graph_construction} and formulate our graph-augmented summarization framework in \S~\ref{sec:model}. In \S~\ref{sec:reinforce}, we introduce reinforcement learning with cloze reward. Experiments and results are presented in \S~\ref{sec:experiments} and \S~\ref{sec:result}. Finally, we conclude in \S~\ref{sec:conclude}.

\section{Related Work}
\label{sec:related}

\noindent \textbf{Graph-Augmented Summarization and Generation.} 
Graph structures have long been used for extractive summarization, such as in Textrank~\cite{mihalcea2004textrank} and Lexrank~\cite{erkan2004lexrank}. For neural models, \newcite{tan-etal-2017-abstractive} design graph-based attention to identify important sentences. For generating abstractive summaries, \newcite{fernandes2018structured} enhance a sequence-based encoder with graph neural networks (GNNs) to consider token-level entity types, however, entity interactions are largely ignored. 
On multi-document summarization, \newcite{fan-etal-2019-using} demonstrate the usefulness of encoding a linearized knowledge graph from OpenIE outputs. 
In this work, we design a graph encoder, which improves upon Graph Attention Networks (GATs)~\cite{velickovic2018graph}, to capture the global context in a more effective manner.

Also related is the graph-to-sequence framework that has been adopted for text generation~\cite{song-etal-2018-graph}. 
Both Gated Graph Neural Networks (GGNNs)~\cite{beck-etal-2018-graph} and Graph Convolutional Networks (GCNs) ~\cite{damonte2019structural} are shown to be effective in generating sentences from AMR graphs. Since Graph Attention Networks can better handle sparse graphs, they are used by \newcite{koncel-kedziorski-etal-2019-text} with a transformer model to create scientific paper abstracts from knowledge graphs. Here we use graphs {\it in addition to} document encoder, both carrying complementary information for summarization.


\smallskip
\noindent \textbf{Reinforcement Learning and QA Reward for Abstractive Summarization.} 
As pointed out by~\newcite{DBLP:journals/corr/RanzatoCAZ15}, word-level maximum likelihood training brings the problem of exposure bias. Recent work utilizes reinforcement learning to directly optimize the model to maximize the informativeness of summaries by using different forms of ROUGE scores~\cite{paulus2018a,chen2018fast,sharma-etal-2019-entity}. However, ROUGE does not always distinguish good summaries from bad ones~\cite{novikova2017we}, and ignores entity interactions.

Since question answering (QA) has been used for summary evaluation~\cite{narayan2018don}, and is shown to correlate with human judgment of summaries qualities~\cite{eyal-etal-2019-question}, QA-based rewards have been studied for summarization model training. 
\newcite{arumae-liu-2019-guiding} demonstrate that using fill-in-the-blank questions by removing entities or root words leads to improved content selection. 
\newcite{scialom-etal-2019-answers} consider a similar setup, but use both F1 score and QA system confidence as rewards in abstractive summarization. 
Previous work, however, mainly focuses on single entities or words in human-written summaries, thereby losing contexts and relations.
Moreover, fill-in-the-blank questions by prior work give credits only when the answers exactly match the ground-truths, thus causing inaccuracies for rephrased answers and discouraging abstract content generation. 
In contrast, we design a semantic-driven cloze reward by measuring how well a QA system can address {\it multiple choice} cloze questions which better {\it encode entity interactions} and {\it handle paraphrased answers}.

\section{Knowledge Graph Construction}
\label{sec:graph_construction}

To construct a knowledge graph from an input document, we utilize Stanford CoreNLP~\cite{manning-EtAl:2014:P14-5} to first obtain outputs from coreference resolution and open information extraction (OpenIE) models~\cite{angeli-etal-2015-leveraging}. Note that we do not conduct global entity linking across documents.  
Next, we take the $\langle$subject, predicate, object$\rangle$ triples extracted by OpenIE and remove any triple whose argument (subject or object) has more than $10$ words. If two triples differ only by one argument, and the arguments overlap, we keep the longer triple. 

We begin constructing the graph by treating subjects and objects as nodes connected by directed edges, with predicates as attributes. We further collapse coreferential mentions of the same entity into one node. With this, we can localize salient content related to each entity as well as make connections of spread-out entities through graph paths.


\begin{figure}[t]
    \centering
    \includegraphics[width=\columnwidth,trim=0 0 0cm 0, clip]{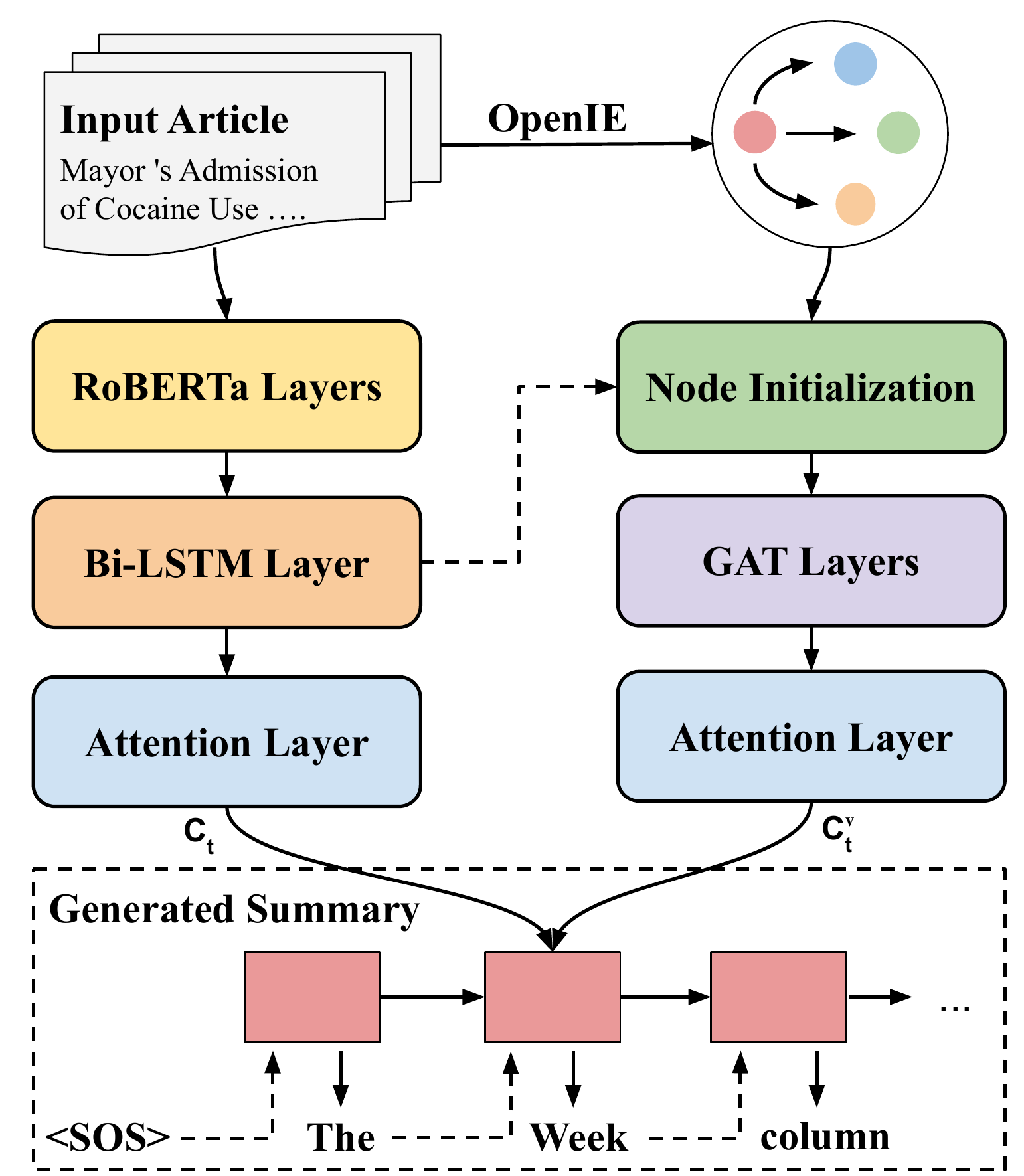}
    \captionof{figure}{ 
    Our \textsc{ASGARD} framework with document-level graph encoding. Summary is generated by attending to both the graph and the input document. 
    }
    \label{fig:model}
\end{figure}

\section{Summarization Model} 
\label{sec:model}

In this section, we describe our graph-augmented abstractive summarization framework, as displayed in Fig.~\ref{fig:model}. 
Our model takes as input a document, represented as a sequence of tokens $\mathbf{x}$ = $\{x_k\}$, and a knowledge graph $G$ consisting of nodes $\{v_i\}$. $\mathbf{x}$ and $G$ are separately consumed by a document encoder and a graph encoder, as presented in \S~\ref{sec:encoders}. Importantly, we present two types of graphs: \textsc{DocGraph}, focusing on the global context, and \textsc{SegGraph}, which additionally captures topic shift. 
The summary decoder then generates an abstractive summary by attending to both the document and the graph (\S~\ref{sec:decoder}). 
In \S~\ref{sec:objectives}, we formulate a maximum likelihood training objective which leverages the detection of salient nodes in the graph.

\subsection{Encoders}
\label{sec:encoders}

\smallskip
\noindent \textbf{Document Encoder.}
We first feed input $\mathbf{x}$ to RoBERTa~\cite{DBLP:journals/corr/abs-1907-11692} and take the last layer output as token embeddings. We then employ a single-layer bidirectional LSTM (BiLSTM) over token embeddings, producing encoder hidden states $\mathbf{h}_k$ at time step $k$. 

\smallskip
\noindent \textbf{Graph Encoder.} 
Built on the graph constructed in \S~\ref{sec:graph_construction}, we create nodes for predicates as done in previous graph-to-sequence work~\cite{beck-etal-2018-graph} to reduce model parameters. Directed, unlabeled edges are added from subject to predicate, and from predicate to object. We further add reverse edges and self-loops to enhance the information flow, and this forms the graph $G$. 

\smallskip
\noindent \textit{Node Initialization.} 
Each node often contains multiple mentions of an entity; we thus initialize node representation $\mathbf{v}_i$ by using the average embedding of its tokens. 
We leverage document encoder hidden states $\mathbf{h}_k$ as the contextual representation of tokens. Number of mentions in the node is added as an extra encoding to $\mathbf{v}_i$, to signify entity salience. 

\smallskip
\noindent \textit{Contextualized Node Encoding.}  
Our graph encoder improves upon Graph Attention Networks (GATs)~\cite{velickovic2018graph} by adding residual connections between layers as discussed in~\newcite{koncel-kedziorski-etal-2019-text}. Each node $\mathbf{v}_i$ is represented by a weighted average of its neighbors:  

\vspace{-3mm}
{\fontsize{10}{11}\selectfont
\begin{align}
\hat{\mathbf{v}}_i &= \mathbf{v}_i + \mathbin\Vert_{n=1}^{N} \sum_{v_j \in \mathcal{N} (v_i)} \alpha_{i,j}^n \mathbf{W}_{0,n} \mathbf{v}_j  \\
\alpha_{i,j}^n &=  \mathrm{softmax}((\mathbf{W}_{1,n}\mathbf{v}_i)^T(\mathbf{W}_{2,n}\mathbf{v}_j)) 
\end{align}
}
%
where $\Vert_{n=1}^{N}$ denotes the concatenation of $N$ heads, each producing a vector of the same dimension as $\mathbf{v}_i$. We use $N=4$ in our experiments with two layers of GATs. 
$\mathcal{N} (v_i)$ denotes the neighbors of $v_i$ in graph $G$. 
$\mathbf{W}_{\ast}$ are trainable parameters.  

The graph encoder described above encodes document-level global context by merging entity mentions throughout the document and capturing their interactions with graph paths. It is henceforth denoted as \textbf{\textsc{DocGragh}}. 

\smallskip
\noindent \textbf{Encoder Extension to Capture Topic Shift (\textsc{SegGragh}).} 
Modeling topic transitions and recurrences enables the identification of notable content, thus benefiting summarization~\cite{barzilay2004catching}. Since paragraphs naturally divide a document into different topic segments, we extend DocGragh by first encoding each paragraph as a subgraph $G_p$ (for the $p$-th paragraph) using the same graph encoder, and then connecting all subgraphs with a BiLSTM. 
If two nodes in separate subgraphs refer to the same entity, they are initialized with the same embedding (as in the first occurrence). 
Concretely, we first apply max-pooling over all nodes in subgraph $G_p$ from the outputs of the final GAT layer; the max-pooling results are then used as inputs for a BiLSTM to produce the final subgraph representation $\mathbf{h}_p^g$ for $G_p$. 




\subsection{Summary Decoder}
\label{sec:decoder}

Our summary decoder uses a single-layer unidirectional LSTM with a hidden state $\mathbf{s}_t$ at step $t$; it generates summary tokens recurrently by jointly attending to the input document and the graph. 

\smallskip
\noindent \textbf{Attending the Graph.}
At each decoding step $t$, we compute a graph context vector $\mathbf{c}^v_t$ with the attention mechanism \cite{bahdanau2014neural}: 

\vspace{-3mm}
{\fontsize{10}{11}\selectfont
\begin{align} \label{eq:battn}
\mathbf{c}^v_t &= \sum_{i} a^v_{i,t} \hat{\mathbf{v}}_i \\
a^v_{i, t} &=  \mathrm{softmax}(\mathbf{u}_0^T  \tanh(\mathbf{W}_{3} \mathbf{s}_t + \mathbf{W}_{4}  \hat{\mathbf{v}}_i)) 
\end{align}} 
where $\mathbf{u}_{\ast}$ are also trainable parameters. We omit bias terms for simplicity.

\smallskip
\noindent \textbf{Attending the Document.} 
Similarly, the document context $\mathbf{c}_t$ is computed over input tokens by additionally considering the graph context $\mathbf{c}^v_t$: 

\vspace{-3mm}
{\fontsize{10}{11}\selectfont
\begin{flalign}
\mathbf{c}_t &= \sum_{k} a_{k,t} \mathbf{h}_k \\
\notag a_{k, t} &= \mathrm{softmax}(\\ & \mathbf{u}_1^T
\tanh(\mathbf{W}_{5} \mathbf{s}_t + \mathbf{W}_{6} \mathbf{h}_k  +  \mathbf{W}_{7} \mathbf{c}^v_t))
\end{flalign}} 
\vspace{-5mm}

\noindent \textbf{Token Prediction.} 
Graph and document context vectors, treated as salient content summarized from both sources, are concatenated with the decoder hidden state $\mathbf{s}_t$ to produce the vocabulary distribution $P_{vocab}$: 

\vspace{-3mm}
 {\fontsize{10}{11}\selectfont
 \begin{flalign}
 P_{vocab} &= \mathrm{softmax}(\mathbf{W}_{out}[\mathbf{s}_t|\mathbf{c}_t|\mathbf{c}^v_t])
 \end{flalign}
 }
\vspace{-5mm}

We use {\it weight-sharing} between the input embedding matrix and the matrix $\mathbf{W}_{out}$ to allow reusing linguistic knowledge as proposed by \newcite{paulus2018a}. We further add a copy mechanism similar to \newcite{see-etal-2017-get}, with copy probability as:

\vspace{-3mm}
{\fontsize{10}{11}\selectfont
 \begin{flalign}
  P_{copy} &= \sigma(\mathbf{W}_{copy}[\mathbf{s}_t|\mathbf{c}_t|\mathbf{c}^v_t|\mathbf{y}_{t-1}])
 \end{flalign}}
where $\mathbf{y}_{t-1}$ denotes the embedding for the token predicted at step $t-1$. 
 


 %
 
 

\smallskip
\noindent \textbf{Modified Hierarchical Attention for SegGraph.} 
As mentioned in \S~\ref{sec:encoders}, SegGraph captures content salience by modeling topic shift across paragraphs. We thus seek to leverage paragraph-level importance to redistribute the node attentions, e.g., giving more attentions to nodes in important paragraphs. 
In particular, we utilize hierarchical attention~\cite{hsu-etal-2018-unified}, where we first calculate attention $\mathbf{a}^g_t$ over subgraphs as done in Eq.~\ref{eq:battn} by replacing $\hat{\mathbf{v}}_i$ with subgraph representation $\mathbf{h}^g_p$. 

We then combine subgraph attentions $\mathbf{a}^g_t$ with the previously calculated attentions $\mathbf{a}^v_t$ for nodes in the subgraph using scalar multiplication and renormalization over all nodes in input. This results in the new attention weights $\hat{\mathbf{a}}^v_t$, which are used to obtain graph context vector $\mathbf{c}^v_t$ as done in Eq.~\ref{eq:battn} for SegGraph.




\subsection{Training Objectives}
\label{sec:objectives}

We first consider a maximum likelihood (ML) training objective that minimizes the following loss: 

\vspace{-3mm}
{\fontsize{10}{11}\selectfont
\begin{flalign}
\mathcal{L}_{\rm seq} &= - \frac{1}{\vert D \vert} \sum_{(\mathbf{y}, \mathbf{x}) \in D} {\log{p(\mathbf{y}\, | \,\mathbf{x};\theta)}}
\end{flalign} 
\label{eq:lmloss}
}  
where $\mathbf{x}$ are documents and $\mathbf{y}$ are references from the training set $D$, and $\theta$ are model parameters. 

\smallskip
\noindent \textbf{Node Salience Labeling.} 
In addition to modeling local characteristics of nodes, we further enhance the model by adding an objective to label node salience, e.g., whether the entities in a node are mentioned in the reference summaries.
We introduce a soft mask layer over each node before it is passed into the graph encoder, to signify its salience. This layer, serving as an information gate, predicts a real number $m_i$ in $[0,1]$ for each node $\mathbf{v}_i$ and multiplies to itself, i.e. $m_i \mathbf{v}_i$. For node $\mathbf{v}_i$, the mask is calculated as $\hat{m}_i = \textrm{sigmoid}(\mathbf{u}_2 \mathbf{v}_i)$. 
During training, the gold-standard mask $m_i$ for a node is set to $1$ if it contains at least one content word in the reference summary; otherwise, $0$. We add the following objective for all nodes in the dataset $D$: 


\vspace{-3mm}
{
\fontsize{10}{11}\selectfont
\begin{flalign}
\notag \mathcal{L}_{mask} = -\frac{1}{N_v} \sum_{v_i \in D} & {m}_{i} \log(\hat{m}_{i})  + \\ & (1 - {m}_{i}) \log(1 - \hat{m}_{i})
\end{flalign}
} 
where $N_v$ represents the number of nodes in the dataset. 
Finally, the ML training objective takes the following form: $\mathcal{L}_{\rm ml} = \mathcal{L}_{\rm mask} +  \mathcal{L}_{\rm seq}$.

\section{Reinforcement Learning with Cloze}
\label{sec:reinforce}

After maximum likelihood training with $\mathcal{L}_{\rm ml}$, we further design a {\it multiple choice cloze reward} in a second-stage reinforcement learning (RL), leading the model to generate more faithful and informative summaries.

For RL, we use a {\it self-critical policy gradient} algorithm~\cite{rennie2017self}. During training, two summaries are generated: 
first, a summary $\mathbf{y}^s$, {\it sampling} tokens based on the probability distribution $p(\mathbf{y}^s | \,\mathbf{x};\theta)$ at each decoding step; 
and second, a \textit{baseline} summary $\hat{\mathbf{y}}$ which greedily selects the tokens of the highest probability at each step. 
The objective of RL is defined based on the rewards of the two summaries, $R(\mathbf{y}^s)$ and $R(\hat{\mathbf{y}})$, as follows:

\vspace{-4mm}
{\fontsize{9}{8}\selectfont
\begin{flalign} \label{eq:rlrloss}
\notag \mathcal{L}_{\rm rl} &=  \\ & -\frac{1}{\vert D \vert}  \sum_{(\mathbf{y}^s, \mathbf{x}) \in D} (R(\mathbf{y}^s) - R(\hat{\mathbf{y}}))\log{p(\mathbf{y}^s | \,\mathbf{x};\theta)}
\end{flalign}
} 
%

Our {\bf reward} function uses the combination of ROUGE and the multiple choice cloze score introduced below, i.e., {$R(\mathbf{y})=R_{rouge}({\mathbf{y}}) + \gamma_{cloze} R_{cloze}$}. 
For ROUGE, it considers F1 scores of ROUGE-1, ROUGE-2, and ROUGE-L calculated against the reference summary, and takes the form of 
{$R_{rouge}({\mathbf{y}})= \gamma_{1} R_{rouge-1}(\mathbf{y}) + \gamma_{2} R_{rouge-2}(\mathbf{y}) +  (1-\gamma_{1}-\gamma_{2}) R_{rouge-L}(\mathbf{y})$}.

\begin{figure}[t]
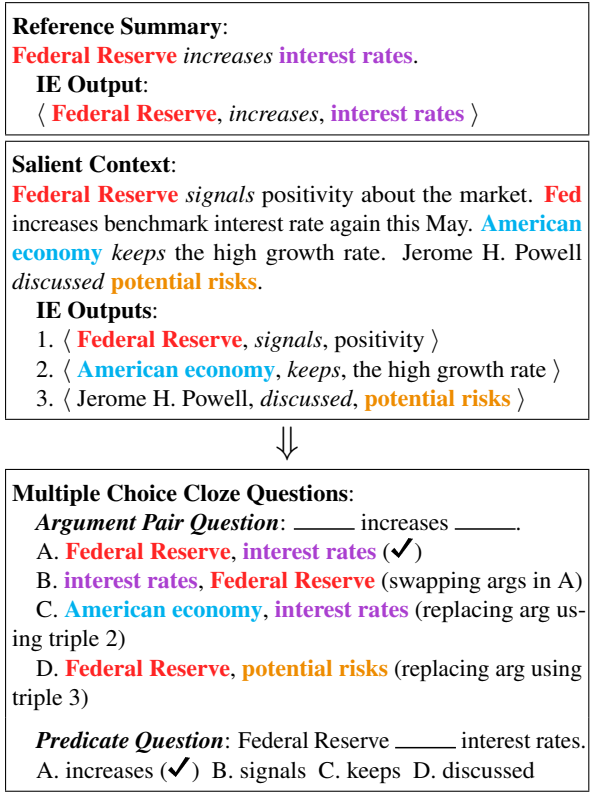

    \def\arraystretch{1.5}
    \bgroup
    \def\arraystretch{1.5}
	\fontsize{9}{11}\selectfont
	\setlength{\tabcolsep}{0.8mm}
	\begin{tabular}{|p{75mm}|}
	\hline
	\textbf{Reference Summary}: 
    
    \textcolor{IndianRed1}{\textbf{Federal Reserve}} \textit{increases} \textcolor{MediumOrchid}{\textbf{interest rates}}.
    
	\quad \textbf{IE Output}: 
	
    \quad $\langle$ \textcolor{IndianRed1}{\textbf{Federal Reserve}}, \textit{increases}, \textcolor{MediumOrchid}{\textbf{interest rates}} $\rangle$ \\
    \hline
    \hline
    
    \textbf{Salient Context}: 
    
    \textcolor{IndianRed1}{\textbf{Federal Reserve}} \textit{signals} positivity about the market. \textcolor{IndianRed1}{\textbf{Fed}} increases benchmark interest rate again this May. 
    \textcolor{DeepSkyBlue1}{\textbf{American economy}} \textit{keeps} the high growth rate. Jerome H. Powell \textit{discussed} \textcolor{Orange}{\textbf{potential risks}}.
    
    \quad \textbf{IE Outputs}:
    
    \quad 1. $\langle$ \textcolor{IndianRed1}{\textbf{Federal Reserve}}, \textit{signals}, positivity $\rangle$
    
    
    \quad 2. $\langle$ \textcolor{DeepSkyBlue1}{\textbf{American economy}}, \textit{keeps}, the high growth rate $\rangle$ 
    
    \quad 3. $\langle$ Jerome H. Powell, \textit{discussed}, \textcolor{Orange}{\textbf{potential risks}} $\rangle$ \\
    \hline
    \end{tabular}
    
    \begin{tabular}{c}
    \hspace{34mm} {\Large $\Downarrow$}\\
    \end{tabular}
    
    \begin{tabular}{|p{75mm}|}
    \hline
    \textbf{Multiple Choice Cloze Questions}:
    
    \quad \textit{\textbf{Argument Pair Question}}: $\rule{0.8cm}{0.2mm}$ increases $\rule{0.8cm}{0.2mm}$. 
    
    \quad A. \textcolor{IndianRed1}{\textbf{Federal Reserve}}, \textcolor{MediumOrchid}{\textbf{interest rates}} (\checkmark)
    
    \quad B. \textcolor{MediumOrchid}{\textbf{interest rates}}, \textcolor{IndianRed1}{\textbf{Federal Reserve}} ({swapping} args in A)
    
    \quad C. \textcolor{DeepSkyBlue1}{\textbf{American economy}}, \textcolor{MediumOrchid}{\textbf{interest rates}} ({replacing} arg using triple 2)
    
    \quad D. \textcolor{IndianRed1}{\textbf{Federal Reserve}}, \textcolor{Orange}{\textbf{potential risks}} ({replacing} arg using triple 3)\\
    
    \quad \textit{\textbf{Predicate Question}}: Federal Reserve $\rule{0.8cm}{0.2mm}$ interest rates.
    
    \quad A. increases (\checkmark)
    \ B. signals   
    \ C. keeps   
    \ D. discussed \\
    \hline
	
	\end{tabular}
\caption{Sample construction of multiple choice cloze questions and candidate answers from reference summary and salient context. Arguments and predicates in candidate answers are color-coded and italicized. 
} 
\label{fig:clozeqa}
\egroup
\end{figure}

\smallskip
\noindent \textbf{Multiple Choice Cloze Reward.} 
Here, we present a novel multiple choice cloze reward to work with our knowledge graph and guide the summarization model towards improved awareness of entity interactions. 
We treat the system-generated summary as \textbf{context}. We provide a set of \textbf{questions} automatically constructed from the corresponding reference summary written by a human. 
We separately train a question answering (QA) model to address the questions by reading the context. Intuitively, if the system summary shares salient information with the reference, the QA model will assign the correct answers with high probability. We decide to use the average probability of the correct answers as our \textbf{cloze reward}. 
Below, we give details on how to construct the questions and candidate answers with examples shown in Fig.~\ref{fig:clozeqa}. 

\smallskip
\noindent \textit{Question Construction.} 
We run the OpenIE tool on human-written summaries, retaining triples with arguments not longer than $5$ words. For each triple of $\langle$subject, predicate, object$\rangle$, we create two types of questions: 
(1) \textbf{argument pair questions}, by removing the subject and object, and
(2) \textbf{predicate questions}, by removing the predicate.

\smallskip
\noindent \textit{Candidate Answer Construction.} 
Because fill-in-the-blank style cloze may incorrectly penalize QA systems with answers paraphrased from the ground-truth, we opt for a multiple choice cloze. We construct three {\bf candidate answers} in addition to the gold-standard from the {\bf salient context}, which are summary-worthy sentences selected from the input. Specifically, we use greedy search to select the best combination of sentences that maximizes ROUGE-2 F1 with reference to human summary. We further include a sentence in the salient context if it has a ROUGE-L recall greater than $0.6$ when compared with any sentence in the reference.

We first select OpenIE triples from the salient context and filter out those that have any overlapping content word with the correct answer. For {\it argument pair questions}, we create one candidate answer by swapping the subject and the object (e.g. candidate B as in Fig.~\ref{fig:clozeqa}) and two candidates by replacing the subject or the object with another argument of the same role extracted from the salient context (e.g. candidates C and D). 
If not enough answers are created, we further consider randomly selecting sentences from the input. 
For {\it predicate questions}, we use predicates in other triples from the context as candidate answers. Among all candidates, we select the three that are able to construct the most fluent questions using perplexity predicted by BERT~\cite{devlin-etal-2019-bert}. 

In case reference summaries do not yield OpenIE triples, we create additional entity pair questions. We remove two co-occurring entities from the summary and create three candidate answers in the same way as described above. 



\smallskip
\noindent \textbf{QA Model.} 
We fine-tune RoBERTa~\cite{DBLP:journals/corr/abs-1907-11692} to build our QA model. We use the salient context described above as the context for training. We then concatenate the context, the question, and each of the four candidate answers, and pass the final [CLS] representation through a fully-connected layer, from which the answer is predicted. 

\section{Experimental Setups}
\label{sec:experiments}

\noindent \textbf{Datasets.} 
We experiment with two popular summarization datasets with summaries containing multiple sentences: the New York Times annotated corpus (NYT)~\cite{sandhaus2008new} and the CNN/Daily Mail dataset (CNN/DM)~\cite{hermann2015teaching}. We follow the preprocessing steps and experimental setups from prior work~\cite{paulus2018a,see-etal-2017-get} for both datasets. For NYT, the training, validation, and test sets contain $588,909$, $32,716$, and $32,703$ samples. For CNN/DM, the numbers are $287,188$, $13,367$, and $11,490$. 

To train our cloze QA model for NYT, we construct $1,414,336$ question-answer pairs from human-written summaries in the training set based on the method described in \S~\ref{sec:reinforce}. 
On CNN/DM, we collect $1,361,175$ question-answer samples from the training set. 
For both datasets, we set aside $20,000$ samples as a validation set and $20,000$ samples as a test set. Our QA model achieves an accuracy of $97\%$ on NYT and $95\%$ on CNN. 
%

\smallskip
\noindent \textbf{Training Details and Parameters.} 
We use the base version of RoBERTa model to extract token features for all experiments. 
We truncate input articles to $1024$ (NYT) and $512$ (CNN/DM) BPEs. 
We employ LSTM models with $256$-dimensional hidden states for the document encoder ($128$ each direction) and the decoder. For the residual connection of the graph encoder, we use $4$ heads, each with a dimension of $72$. 
For DocGraph training and inference, we prune isolated graphs with fewer than three nodes to increase robustness and reduce redundancy. 
We set $\gamma_{1} = 0$, $\gamma_{2} = 0.75$ on NYT and $\gamma_{1} = 0.33$, $\gamma_{2} = 0.33$ on CNN/DM after tuning on the validation set. 
For both datasets, we set $\gamma_{cloze} = 0.05$. More details about parameters and graph statistics are in the Appendices.

\smallskip
\noindent \textbf{Baselines and Comparisons.} 
For both datasets, we include an extractive baseline \textsc{Lead-3}. We further add the following abstractive models for comparison: (1) a pointer-generator model with coverage~\cite{see-etal-2017-get} (\textsc{PointGen+cov}); (2) a deep reinforcement learning-based model~\cite{paulus2018a} (\textsc{DeepReinforce}); (3) a bottom-up model~\cite{gehrmann-etal-2018-bottom} (\textsc{BottomUp}); (4) a deep communicating agents-based summarization model~\cite{celikyilmaz-etal-2018-deep} (\textsc{DCA}). 
We also report results by fine-tuning BART model~\cite{lewis2019bart}. In \newcite{lewis2019bart}, fine-tuning is only performed on CNN/Daily Mail. We apply the same method for NYT.

For NYT, we add results by \textsc{SENECA} model~\cite{sharma-etal-2019-entity} from our prior work, which previously achieved the best ROUGE-2. 

On CNN/Daily Mail, we include comparisons of a two-stage fine-tuned model (first on an extractor, then on an abstractor) with BERT~\cite{liu-lapata-2019-text} (\textsc{BertSumExtAbs}), and a unified pretrained language model for generation~\cite{NIPS2019_9464} (\textsc{UniLM}).




In addition to 
\textsc{ASGARD-doc} and \textsc{ASGARD-seg}, which are trained with an ML objective, we report results trained with ROUGE as the reward ($R_{rouge}$), and with an additional cloze reward ($R_{cloze}$). 
Lastly, we consider a variant \textsc{NoGraph} by ablating the graph encoder. 
\section{Results}
\label{sec:result}

\subsection{Automatic Evaluation}

\begin{table}[t]
\centering
\fontsize{9}{11}\selectfont
\setlength{\tabcolsep}{1.0mm}
\begin{tabular}{@{}lccc@{}}
\toprule
\textbf{System} & \textbf{ROUGE-1} & \textbf{ROUGE-2} & \textbf{ROUGE-L}  \\\midrule
\textsc{Lead-3} & 32.59	& 16.49	& 29.17\\ \hline
\textsc{PointGen+cov} \hfill &  41.06  & 25.71 & 37.28     \\
\textsc{DeepReinforce} &  47.03 &30.72 &	43.10   \\ 
\textsc{BottomUp} & 47.38 &	31.23 &	41.81  \\
\textsc{DCA} & 48.08 & 31.19 & 42.33  \\
\textsc{SENECA} & 47.94 & 31.77 & 44.34	 \\
\textsc{BART} & \textbf{53.25} & \textbf{36.61} & \textbf{48.78}  \\
\hline
{\bf Our Models}& 	& 	& \\
\textsc{NoGraph} & 47.15 & 32.02 & 43.65  \\
\quad $+R_{rouge}$ & 49.17 & 33.19 & 46.44 \\
\textsc{ASGARD-doc} & 49.51 & 33.82 & 45.72  \\
\quad $+R_{rouge}$ & 50.18 & 33.91 & 46.84    \\
\quad $+R_{rouge}+R_{cloze}$ & 50.59 & 33.98 &  48.24   \\
\textsc{ASGARD-seg} & 49.54 & 33.84 & 45.75   \\ 
\quad $+R_{rouge}$ & 50.47 & 33.95 & 47.43   \\
\quad $+R_{rouge}+R_{cloze}$ & \textit{51.29} & \textit{34.97} & \textit{48.26}   \\
\bottomrule
\end{tabular}
\caption{Automatic evaluation with ROUGE on New York Times. Best results are in \textbf{boldface}. Best of our models are in \textit{italics}. \textsc{ASGARD-seg}+$R_{rouge}$+$R_{cloze}$ yields significantly higher scores than
our other models with approximate randomization test ($p<0.0005$). 
}
\label{tab:nyt-rouge}
\vspace{-7mm}
\end{table}

\begin{table}[ht]
\centering
\fontsize{9}{11}\selectfont
\setlength{\tabcolsep}{1.0mm}
\begin{tabular}{@{}lccc@{}}
\toprule
\textbf{System} & \textbf{ROUGE-1} & \textbf{ROUGE-2} & \textbf{ROUGE-L}  \\\midrule
\textsc{Lead-3} & 40.23	& 17.52	& 36.34  \\ \hline
\textsc{PointGen+cov} & 39.53 &	17.28 &	36.38   \\ 
\textsc{DeepReinforce} &  41.16 & 15.75 & 39.08  \\
\textsc{BottomUp} & 41.22 & 18.68 & 38.34   \\
\textsc{DCA} & 41.69 & 19.47 & 37.92  \\
\textsc{BERTSumExtAbs} & 42.13 & 19.60 & 39.18\\
\textsc{UniLM} & 43.33 & 20.21 & 40.51  \\
\textsc{BART} & \textbf{44.16} & \textbf{21.28} & \textbf{40.90}  \\
\hline
{\bf Our Models}& 	& 	& \\
\textsc{NoGraph} & 39.55 & 17.89 & 36.75 \\
\quad $+R_{rouge}$ & 41.37 & 17.63 & 37.99  \\
\textsc{ASGARD-doc} & 40.38 & 18.40 & 37.51 \\
\quad $+R_{rouge}$ & 43.10 & 17.58 & 39.41  \\
\quad $+R_{rouge}+R_{cloze}$ & \textit{43.93} & \textit{20.37} & \textit{40.48}   \\
\textsc{ASGARD-seg} & 40.09 & 18.30 & 37.30 \\
\quad $+R_{rouge}$ & 42.94 & 17.93 & 39.36  \\
\quad $+R_{rouge}+R_{cloze}$ & 43.81 & 20.22 & 40.37   \\
\bottomrule
\end{tabular}
\caption{Automatic evaluation with ROUGE on CNN/Daily Mail. 
Best results of our model variants are in \textit{italics}. Both \textsc{ASGARD-seg}+$R_{rouge}$+$R_{cloze}$ and \textsc{ASGARD-doc}+$R_{rouge}$+$R_{cloze}$ obtain significantly better scores than other model variants ($p<0.0005$).}
\label{tab:cnn-rouge}
\end{table}

\noindent \textbf{Results on NYT.} 
As displayed in Table~\ref{tab:nyt-rouge}, our \textsc{ASGARD-seg} model trained with ROUGE and cloze rewards achieves better ROUGE scores~\cite{Lin:2003:AES:1073445.1073465} than all other comparisons except the fine-tuned \textsc{BART}. However, our \textsc{ASGARD-seg}'s ROUGE-L score is comparable to BART. 
This indicates the effectiveness of our graph-augmented summarization framework.

%
Moreover, both our \textsc{ASGARD-doc} and \textsc{ASGARD-seg} models yield significantly higher ROUGE scores than the variant without the graph encoder (\textsc{NoGraph}). This demonstrates the benefit of using structured representation to encode entity interactions. 
Furthermore, both \textsc{ASGARD-doc} and \textsc{ASGARD-seg} with cloze reward ($R_{cloze}$) obtain significantly higher scores compared to the models trained with ROUGE reward only. This signifies that our multi-choice cloze reward can guide better semantic interpretation of content, leading to the generation of more informative summaries. 
We also find that \textsc{ASGARD-seg} outperforms \textsc{ASGARD-doc}, indicating that \textsc{ASGARD-seg} better captures topic drift through multiple paragraphs.

\smallskip
\noindent \textbf{Results on CNN/DM.} 
We observe similar trends on the CNN/DM articles as shown in Table~\ref{tab:cnn-rouge}. Noticeably, \textsc{ASGARD-doc} trained with the combined ROUGE and cloze reward produces better ROUGE scores than \textsc{BERTSumExtAbs} and \textsc{UniLM}, which are carefully fine-tuned from large pretrained language models, and the numbers are also comparable to the fine-tuned \textsc{BART}.

\begin{figure}[t]
    \centering
    \includegraphics[width=0.95\columnwidth,trim=0cm 0 0.1cm 0, clip]{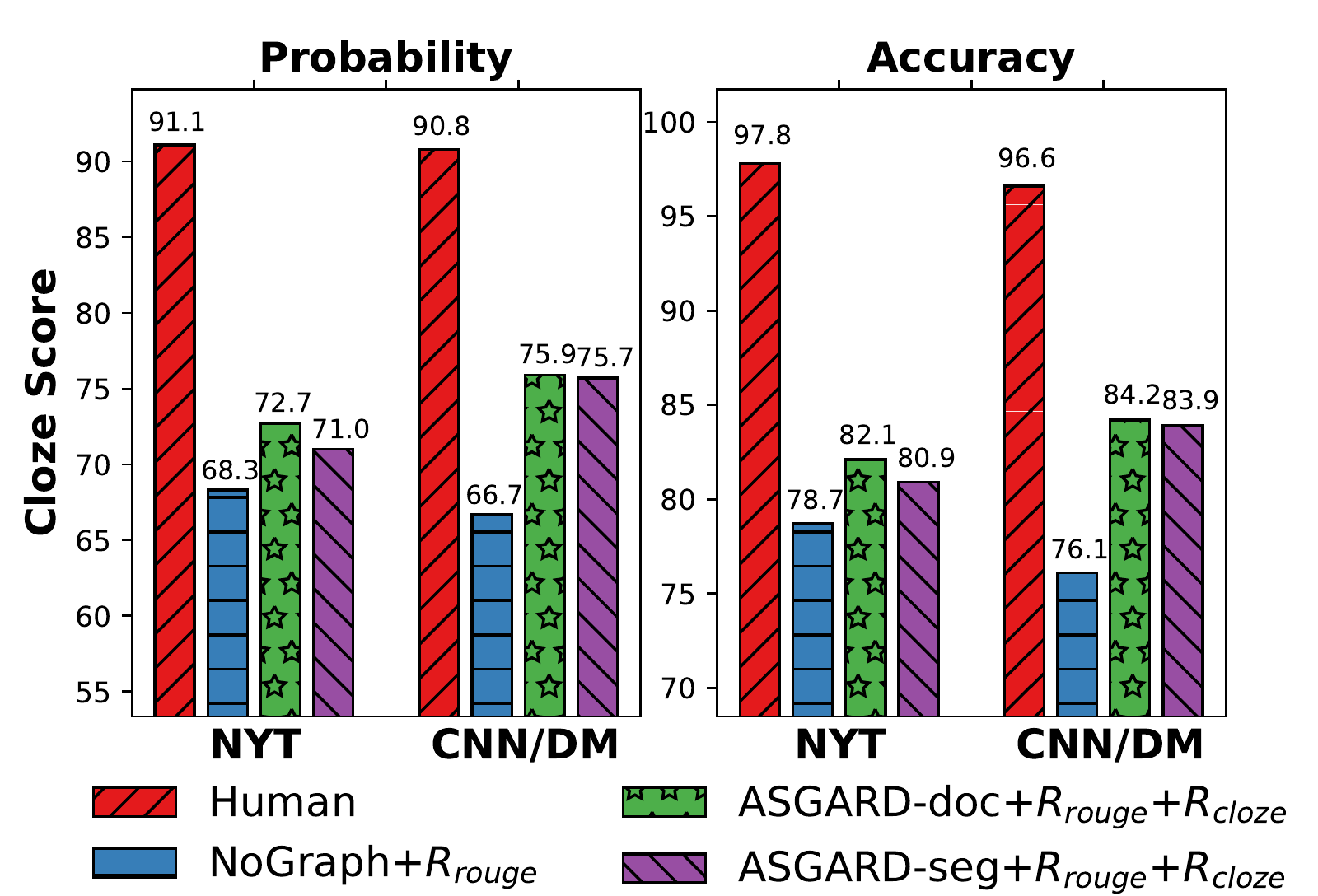}
    \caption{Evaluation with QA model prediction probability and accuracy on our multiple choice cloze test, with higher numbers indicating better summaries.}
    \label{fig:clozescore}
\end{figure}

\smallskip
\noindent \textbf{Evaluation with Cloze Test.} 
We further evaluate model-generated summaries with our proposed cloze test. Here, we report two scores in Fig.~\ref{fig:clozescore}: the average \textbf{probability} of the correct answers output by our QA model, and its prediction \textbf{accuracy}. We first calculate one score per summary, then take the average over all summaries. We can see that our models with graph encoders perform better than the variant without it. 


\subsection{Human Evaluation}
We further conduct human evaluation to analyze the informativeness and fluency of the generated summaries, as well as to investigate the unfaithful errors made by different models. 
We sample $100$ articles from the NYT test set and hire three native or fluent speakers of English to rate summaries generated by our two systems, \textsc{NoGraph}$+R_{rouge}$ and \textsc{ASGARD-seg}$+R_{rouge}+R_{cloze}$, along with outputs by \textsc{BART} and human-written summaries (presented in random order). 
After reading the articles, each judge scores summaries on a Likert scale from 1 (worst) to 5 (best) on \textbf{informativeness}---whether the summary covers important information from the input, and \textbf{fluency}---whether the summary is grammatically correct.

We consider three types of unfaithful errors: (i) \textbf{hallucination error}---creating content not present in the input, 
(ii) \textbf{out-of-context error}---generating facts without including required context or within incorrect context, 
and (iii) \textbf{deletion or substitution error}---mistakenly deleting or substituting subjects, objects, or clauses. 
We ask the annotators to label each type as $1$ for existence of errors, and $0$ otherwise. Detailed guidelines are in the Appendices.  


\begin{table}[t]
\centering
\fontsize{9}{10}\selectfont
 \setlength{\tabcolsep}{0.6mm}
  \centering
    \begin{tabular}{lccccc}
        \toprule
        
        \textbf{System} & {\bf Inf.}$\uparrow$ & {\bf Flu.}$\uparrow$ & \textbf{Hal.}$\downarrow$ & \textbf{Out.}$\downarrow$ & \textbf{Del./Sub.}$\downarrow$ \\
        \midrule
        \textsc{Human} & 4.47 & 4.65 & 21\% & 10\% & 10\%  \\ \cdashline{1-6}
        \textsc{NoGraph $+R_{rouge}$} & 3.94 & 3.65 & \textbf{9}\%$^{\ast}$ & 26\% & 22\%   \\ 
        \textsc{ASGARD-seg} \\
        \hspace{0.5mm} $+R_{rouge}+R_{cloze}$ & 4.12$^{\dagger}$ & 3.77$^{\dagger}$ & 23\% & \textbf{14}\%$^{\dagger}$ & \textbf{9}\%$^{\ast}$ \\
        \textsc{BART} & \textbf{4.44}$^{\ast}$ & \textbf{4.66}$^{\ast}$ & 16\% & 15\% & 12\% \\
        \bottomrule
    \end{tabular}
    \caption{
    Human evaluation on informativeness (Inf.) and fluency (Flu.) (1-to-5), and percentages of unfaithful errors of hallucination (Hal.), out-of-context (Out.) and deletion or substitution (Del./Sub.). 
    $\ast$: significantly different from all other models. 
    $\dagger$: \textsc{ASGARD-seg} is significantly better than \textsc{NoGraph} ($p<0.05$). 
    Inter-rater agreement with Krippendorf's $\alpha$ for all columns: 0.61, 0.70, 0.57, 0.50 and 0.43.
    }
  \label{tab:human-eval}
\end{table}

\begin{figure}[t]
\centering
\fontsize{9}{10}\selectfont
\setlength{\tabcolsep}{1.5mm}{
	\begin{tabular}{p{74mm}}
    \hline
    \textbf{Summary by Human:} 
    
    Family Court in Burlington County, NJ, rules that lesbian couple can list both their names as parents on birth certificate of newborn; \textcolor{blue!75}{\textbf{state attorney general's office drops opposition to move; court ruling negates couple's having to go through adoption proceedings to establish full parental rights for both.}} \\
    \hline
    \textbf{NoGraph$+R_{rouge}$:} 
    
    Lesbian couple in South Jersey wins court approval to have both of their names listed as parents on birth certificate of their newborn. it will no longer oppose such applications \\
    
    \hline
    \textbf{ASGARD-doc$+R_{rouge}+R_{cloze}$:} 
    
    Lesbian couple in South Jersey, won court approval to have both of their names listed as parents on birth certificate of their newborn.
    \textcolor{green!80!blue!80!red!80!black!90}{\textbf{attorney general's office says it will no longer oppose such applications}} \\
    
    \hline
    \textbf{ASGARD-seg$+R_{rouge}+R_{cloze}$:}
    
    Lesbian couple in South Jersey wins court approval to have both of their names listed as parents on birth certificate of newborn \textcolor{green!80!blue!80!red!80!black!90}{\textbf{and attorney general 's office will no longer oppose such applications.  decision stems from Oct 0 ruling by New Jersey Supreme Court holding that same-sex couples are entitled to same legal rights and protections as heterosexual couples}} \\
    
    \hline
    
	\end{tabular}
	}
	\caption{
	Sample summaries for an NYT article. Summaries by our models with the graph encoder are more informative 
	than the variant without it. 
	}
	
\label{fig:sample-outputs}
\end{figure}

From Table~\ref{tab:human-eval}, we can see that our \textsc{ASGARD-seg} model obtains better scores in informativeness and fluency, compared to the variant without the graph encoder. This indicates the effectiveness of leveraging knowledge graph representation. Sample output summaries by our models can be found in Fig.~\ref{fig:sample-outputs}. 
Meanwhile, fine-tuned \textsc{BART} model produces outputs with similar informativeness and fluency of human-constructed summaries, suggesting a future direction of building our model on top of a large-pretrained encoder-decoder model.

For {\bf unfaithful errors}, we report the percentage of errors calculated by majority voting (i.e., more than one annotator vote as incorrect). 
First, we find that our \textsc{ASGARD-seg} model has a comparable error pattern as human summaries. 
Specifically, for out-of-context and deletion or substitution errors, our graph-enhanced model produces significantly fewer mistakes in these categories, compared to the model without graph information. This implies that knowledge graph-enhanced models can improve summary faithfulness. 

Interestingly, human-written summaries are also discerned to contain a nontrivial amount of hallucination errors. After inspection, we find that human tends to leverage world knowledge to include content that is not covered by the articles. For instance, for an article discussing events in ``Boston", the human writer may describe them as happening in ``Massachusetts" in the summary. 



\subsection{Analyzing Automatic Metrics and Summary Errors}  
\begin{figure}[t]
    \centering
    \vspace{-13mm}
    \includegraphics[width=\columnwidth,trim=0.5cm 0 0.1cm 0, clip]{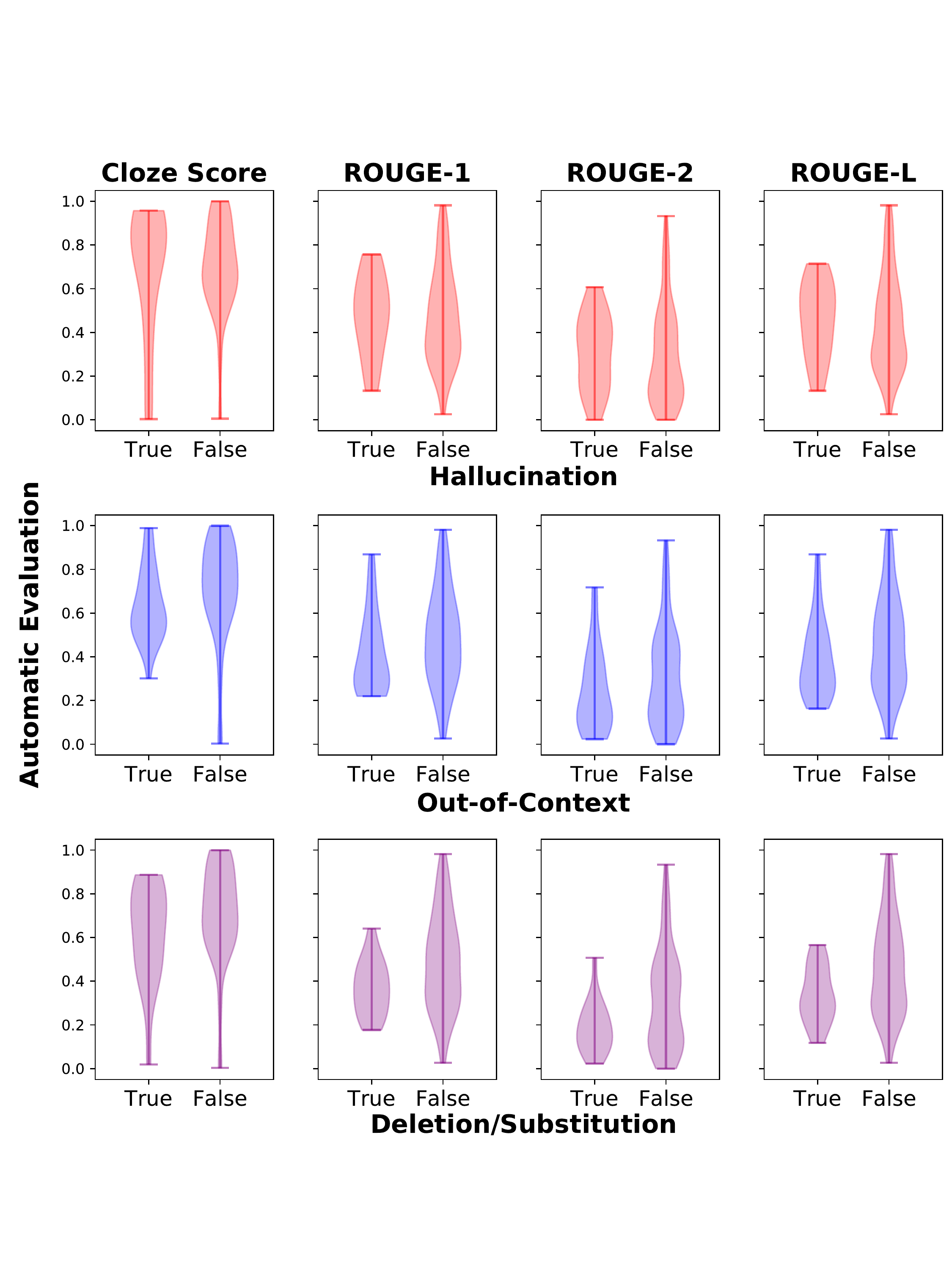}
    \vspace{-17mm}
    \captionof{figure}{
    Distribution of automatic summarization metrics with three types of unfaithful errors. ``True'' indicates summaries \textbf{with} such type of error.
    }
    \label{fig:correlation}
\end{figure}

We further plot the distributions of automatic evaluation scores regarding the three types of unfaithful errors based on majority voting in Fig.~\ref{fig:correlation}. 
First, summaries with out-of-context and deletion or substitution errors receive lower cloze and ROUGE scores overall.

Nevertheless, with regard to hallucination errors, we do not see such pattern; there is even a slightly reversed relation with both cloze scores and ROUGE scores, wherein summaries with more hallucination errors tend to score higher. This echos our previous observation that human summaries can be hallucinatory too, where world knowledge is used for writing the summaries.\footnote{During human evaluation, we do not ask human judges to distinguish the source of hallucination errors, i.e. from world knowledge or out of fabrication, since this requires significant domain knowledge.}

Furthermore, we find a weak correlation between the three variants of ROUGE scores and three types of errors, e.g., the minimum and the maximum values of Pearson's $r$ are $-0.19$ and $0.14$. 
This suggests that new metrics should be designed to better gauge summary quality. We plan to study this direction in future work.

\section{Conclusion}
\label{sec:conclude}
We presented a novel knowledge graph-augmented abstractive summarization framework, along with a novel multiple choice cloze reward for reinforcement learning. Our models capture both local characteristics and global interactions of entities from the input, thus generating summaries of higher quality. In tandem with the graph representation, our cloze reward further improves summary content. Human evaluation further confirms that our graph-augmented models trained with the cloze reward produce more informative summaries and significantly reduces unfaithful errors. 

\section*{Acknowledgements}
\vspace{-1mm}
This research is supported in part by National Science Foundation through Grant IIS-1813341, and by the Office of the Director of National Intelligence (ODNI), Intelligence Advanced Research Projects Activity (IARPA), via contract \# FA8650-17-C-9116. The views and conclusions contained herein are those of the authors and should not be interpreted as necessarily representing the official policies, either expressed or implied, of ODNI, IARPA, or the U.S. Government. The U.S. Government is authorized to reproduce and distribute reprints for governmental purposes notwithstanding any copyright annotation therein. We thank the anonymous reviewers for their suggestions.

\bibliography{reference}
\bibliographystyle{acl_natbib}

\appendix
\section{Appendices}
\subsection{Experiment Details}

\noindent \textbf{Statistics of Knowledge Graphs.} 
We show the statistics of knowledge graphs on two datasets in Table~\ref{tab:kgstat}. On each dataset, we construct a large graph with abundant relations for each article. Note that on CNN/DM we have more arguments but fewer predicates in a document than those on NYT. This indicates CNN/DM has fewer coreferred entities.


\smallskip
\noindent \textbf{Training Details.}
We utilize Adam~\cite{kingma2014adam} with a gradient clipping of $2.0$ and a batch size of $32$ for all models. 
During ML training, a learning rate of $0.001$ is used; during RL stage, it is reduced to $0.0001$~\cite{paulus2018a}. 

We use the base version of BERT model~\cite{devlin-etal-2019-bert} to select candidate answers and we fine-tune the base version of RoBERTa model~\cite{DBLP:journals/corr/abs-1907-11692} to build our QA model. We take pretrained models from~\newcite{Wolf2019HuggingFacesTS}. 

\begin{table}[t]
\centering
\fontsize{9}{11}\selectfont
 \setlength{\tabcolsep}{0.5mm}
  \centering
    \begin{tabular}{lcccccc}
        \toprule
        
        \textbf{Dataset} & {\bf Doc} & \multicolumn{2}{{c}}{\bf \textsc{DocGraph}} & \multicolumn{3}{{c}}{\bf \textsc{SegGraph}} \\
        \cline{3-4}
        \cline{5-7}
         & { \# word}  & { \# Arg.} & { \# Pre.} & { \# Arg.} & { \# Pre.} & { \# Para.}\\
        \midrule
        \textsc{NYT} & 795.9 & 131.6 & 87.3 & 6.40 & 3.74 & 23.5     \\ 
        \textsc{CNN/DM} & 789.9 & 138.1 & 85.2 & 6.30 & 3.57 & 24.2     \\ 
        \bottomrule
    \end{tabular}
    \caption{
    Statistics of NYT and CNN/DM datasets. \# Arg.: number of arguments in each document or paragraph. \# Pre.: number of predicates in each document or paragraph. \# Para.: number of paragraphs in each document. Two datasets have comparable graph size. 
    }
  \label{tab:kgstat}
\end{table}
\subsection{Human Evaluation Guideline}
In our human evaluation, each human annotator is presented with 100 news articles. The annotators are asked to evaluate four summaries (in random order) for each article on two aspects (informativeness and fluency) on a scale of 1 to 5 (1 being very poor and 5 being very good). Furthermore, for unfaithfulness, we define three types of unfaithful errors and ask annotators to label whether summaries contain any type of error. Instructions in Table \ref{tab:human_eval} are given to human judges. 

Here are descriptions of the aspects:

\begin{itemize}
    \item \textbf{Informativeness}: Whether the summary provides enough and necessary content coverage from the input article.
    \item \textbf{Fluency}: Whether the summary is free of obvious grammatically incorrect sentences (e.g., fragments, missing components) that make the text difficult to read.
    \item \textbf{Faithfulness}: Whether the summary accords with the facts expressed in the source. 
\end{itemize}

\newpage
\begin{table*}
	\fontsize{10}{11}\selectfont
    \centering
    \begin{tabular}{lp{120mm}}
         \toprule
         \multicolumn{2}{c}{\textbf{Article: With a Little Extra Cash.}} \\
         \midrule
         & What to do with a bonus? The right thing, of course, is to pay off debts or save it for a time when there are not any bonuses. But in Albany, any financial windfall invites hordes of legislators hungrily seeking ways to spend it. This has already started to happen, with lawmakers eyeballing a projected budgetary surplus of just under \$1 billion -- not all that grand when you consider that the total state budget is in the neighborhood of \$120 billion, but a healthy number nonetheless.  
          \\
          & But one essential part of the equation is different this year: a new governor guarding the state finances. Nobody knows quite yet how Gov. Eliot Spitzer will manage a Legislature that wants to add a lot of its favorite things to his budget before they return it for his approval. One suggestion: Mr. Spitzer should keep his fist as tightly closed as possible, especially on his new school aid formula and his Medicaid adjustments.   \\
          & (....) \\
         \midrule
         \multicolumn{2}{c}{\textbf{Informativeness:}} \\
         \midrule
         \rowcolor{lightgray!30}
          1 & Not relevant to the article \\
          & e.g., \textit{``editorial on gov eliot spitzer 's plan to spend it . of new governor guarding state finances . and to spitzer should keep his fist as tightly closed as possible , especially on new school aid formula and his medicaid adjustments .''} \\
          
          \rowcolor{lightgray!30}
          3 & Relevant, but misses the main point of the article \\
          & e.g., \textit{``editorial on new gov eliot spitzer 's new governor guarding state finances . says spitzer should keep his new school aid formula and his medicaid adjustments''} \\
          \rowcolor{lightgray!30}
          5 &  Successfully captures the main point of the article \\
          & e.g., \textit{``Editorial says New York Gov Eliot Spitzer , faced with projected \$ 0 billion budget surplus , should be tight-fisted and cautious about overspending''} \\
          \midrule
         \multicolumn{2}{c}{\textbf{Fluency:}} \\
         \midrule
         \rowcolor{lightgray!30}
         1 & Summary is full of garbage fragments and is hard to understand  \\
         & e.g., \textit{``of new governor guarding state finances . and to spitzer should keep his fist as tightly closed as possible , to''} \\
         \rowcolor{lightgray!30}
         2 & Summary contains fragments, missing components but has some fluent segments \\
         & e.g., \textit{``editorial on gov eliot spitzer 's plan to spend it .
         of new governor guarding state finances . and to spitzer should keep his fist as tightly closed as possible , especially on new school aid formula and his medicaid adjustments.''} \\
         \rowcolor{lightgray!30}
         3 & Summary contains some grammar errors but is in general fluent\\
         & e.g., \textit{``editorial on any financial windfall invites hordes of legislators hungrily seeking ways to spend it . how gov eliot spitzer will manage legislature that wants to add lot of its favorite to his budget before they return it for his approval .''} \\
         \rowcolor{lightgray!30}
         4 & Summary has relatively minor grammatical errors \\
         & e.g., \textit{``article on in any financial windfall invites hordes of legislators hungrily seeking ways to spend it''} \\
         \rowcolor{lightgray!30}
         5 & Fluent summary \\
         & e.g., \textit{``editorial says new new jersey gov eliot spitzer guarding state finances . says spitzer should keep his new school aid formula and his medicaid adjustments''} \\

          \midrule
         \multicolumn{2}{c}{\textbf{Faithfulness:}} \\
         \midrule
         & We define three types of unfaithful errors. Each type is labeled as ``0'' or ``1'' independently. ``0'' means summary does not make this type of error and ``1'' suggests this type of error occurs.
         Three types of errors are : \\
         \rowcolor{lightgray!30}
         i & \textbf{Hallucination error}: Fabricated content that does not occur in the original article \\
         &  e.g., \textit{``correction of dec 0 about new york column on state budget''} \\
         \rowcolor{lightgray!30}
         ii & \textbf{Out-of-Context error}: Fact occurs in the article, but fails without correct context\\
         &  e.g., \textit{``Editorial says one essential part of the equation is different this year: a new governor guarding the tate finances.''}\\
         \rowcolor{lightgray!30}
        iii & \textbf{Deletion or Substitution error}: Summary contains incorrectly edited, missing elements; or summary incorrectly concatenates elements from different sentences. \\
         & e.g., \textit{``editorial says new new jersey gov eliot spitzer guarding state finances, keeping his new school aid formula adjustments.''} \\
         \bottomrule
    \end{tabular}
    \caption{Sample summaries with explanations on human evaluation aspect scales, and the definition of three types of unfaithful errors.}
    \label{tab:human_eval}
\end{table*}


\end{document}